\documentclass[preprint,12pt]{elsarticle}
\journal{Robotics and Autonomous Systems}

\usepackage{amsmath}
\usepackage{amssymb}
\usepackage{mathtools}

\usepackage{graphicx}
\usepackage{wrapfig}
\usepackage{booktabs}
\usepackage{multirow}
\usepackage{array}
\usepackage{colortbl}
\usepackage{tikz}
\usepackage{adjustbox}
\usepackage{subfigure}
\usepackage{pdflscape}

\usepackage[linesnumbered,ruled,vlined]{algorithm2e}
\usepackage{algpseudocode}

\usepackage{siunitx}
\usepackage{url}
\usepackage{float}
\usepackage{changes} 
\presetkeys{todonotes}{inline,backgroundcolor=yellow}{}
\usepackage{chngcntr}
\usepackage{pifont}
\usepackage{comment}
\usepackage{pdflscape}   

\usepackage{hyperref}   
\usepackage{cleveref}

\Crefname{figure}{Fig.}{Figs.}
\Crefname{equation}{Eq.}{Eqs.}
\Crefname{section}{Sec.}{Secs.}
\crefname{algocf}{Alg.}{Algs.}
\crefname{algocfline}{Alg.}{Algs.}

\algdef{SE}[DOWHILE]{Do}{doWhile}{\algorithmicdo}[1]{\algorithmicwhile\ #1}%

\RestyleAlgo{ruled}

\newcolumntype{L}[1]{>{\raggedright\let\newline\\\arraybackslash\hspace{0pt}}m{#1}}
\newcolumntype{C}[1]{>{\centering\let\newline\\\arraybackslash\hspace{0pt}}m{#1}}
\newcolumntype{R}[1]{>{\raggedleft\let\newline\\\arraybackslash\hspace{0pt}}m{#1}}

\begin{document}
\begin{frontmatter}

\title{\LARGE \bf Can Context Bridge the Reality Gap?\\Sim-to-Real Transfer of Context-Aware Policies}

\author[label1]{Marco Iannotta}
\author[label1]{Yuxuan Yang}
\author[label1]{Johannes A. Stork} 
\author[label2]{Erik Schaffernicht}
\author[label1]{Todor Stoyanov}

\affiliation[label1]{organization={AASS Research Centre, Örebro University},
            city={Örebro},
            country={Sweden}}
\affiliation[label2]{organization={Technology Transfer Center Kitzingen, Technical University of Applied Sciences Würzburg-Schweinfurt},
            city={Kitzingen},
            country={Germany}}

\begin{abstract} Sim-to-real transfer remains a major challenge in reinforcement learning (RL) for robotics, as policies trained in simulation often fail to generalize to the real world due to discrepancies in environment dynamics. Domain Randomization (DR) mitigates this issue by exposing the policy to a wide range of randomized dynamics during training, yet leading to a reduction in performance. While standard approaches typically train policies agnostic to these variations, we investigate whether sim-to-real transfer can be improved by conditioning the policy on an estimate of the dynamics parameters --- referred to as context. To this end, we integrate a context estimation module into a DR-based RL framework and systematically compare SOTA supervision strategies. We evaluate the resulting context-aware policies in both a canonical control benchmark and a real-world pushing task using a Franka Emika Panda robot. Results show that context-aware policies outperform the context-agnostic baseline across all settings, although the best supervision strategy depends on the task.
\end{abstract}


\end{frontmatter}

\section{Introduction}
Reinforcement learning (RL) has achieved significant success in developing robot controllers capable of solving complex tasks~\cite{rl_robotics}.
However, training RL policies directly on physical robots demands extensive interactions with the real environment, making training expensive and dangerous for the robot and its surroundings.
To address these limitations, physics simulation engines are widely used as a safer and more efficient alternative for policy training.
Once a policy has been trained in simulation, it is transferred to the physical robot---a process known as \textit{sim-to-real} transfer~\cite{evol_robotics, rl_robotics, flexible_robotic_grasping}.
Although promising, this paradigm is hindered by the \textit{reality} or \textit{sim-to-real gap}, which refers to the discrepancy between the simulated and real-world environments~\cite{reality_gap, reality_gap_survey}.
This gap often leads to a significant decrease in performance when the policy is deployed in reality, posing a substantial challenge.

Domain Randomization (DR) is a widely adopted approach to mitigate the reality gap~\cite{domain_randomization}.
The core idea is to expose the policy to a broad distribution of simulated environments during training by randomizing various simulation parameters that affect the environment dynamics.
Learning to perform robustly across this diverse range of scenarios makes the policy less reliant on a precise match between the simulated and the real environment.
Standard approaches in DR typically train policies that are \textit{agnostic} to the dynamics parameters randomized in simulation, i.e., the policy is trained to perform robustly under all variations, without explicitly incorporating knowledge of the randomized parameters and relying solely on the observed state. 
Instead, Yu et al.~\cite{online_system_identification} propose coupling the RL policy with an online system identification model (OSI) trained to infer dynamics parameters from recent trajectories.
These inferred parameters condition the policy alongside the observed state, explicitly informing the control policy on the underlying dynamics.
Although this approach has shown promising results in simulation, it has never been validated on a real robot, leaving the effectiveness of dynamics-aware policies in real-world scenarios untested.

\begin{figure}[t!]
\centering
\includegraphics[width=0.6\linewidth,trim={8cm 12cm 8cm 2cm},clip]{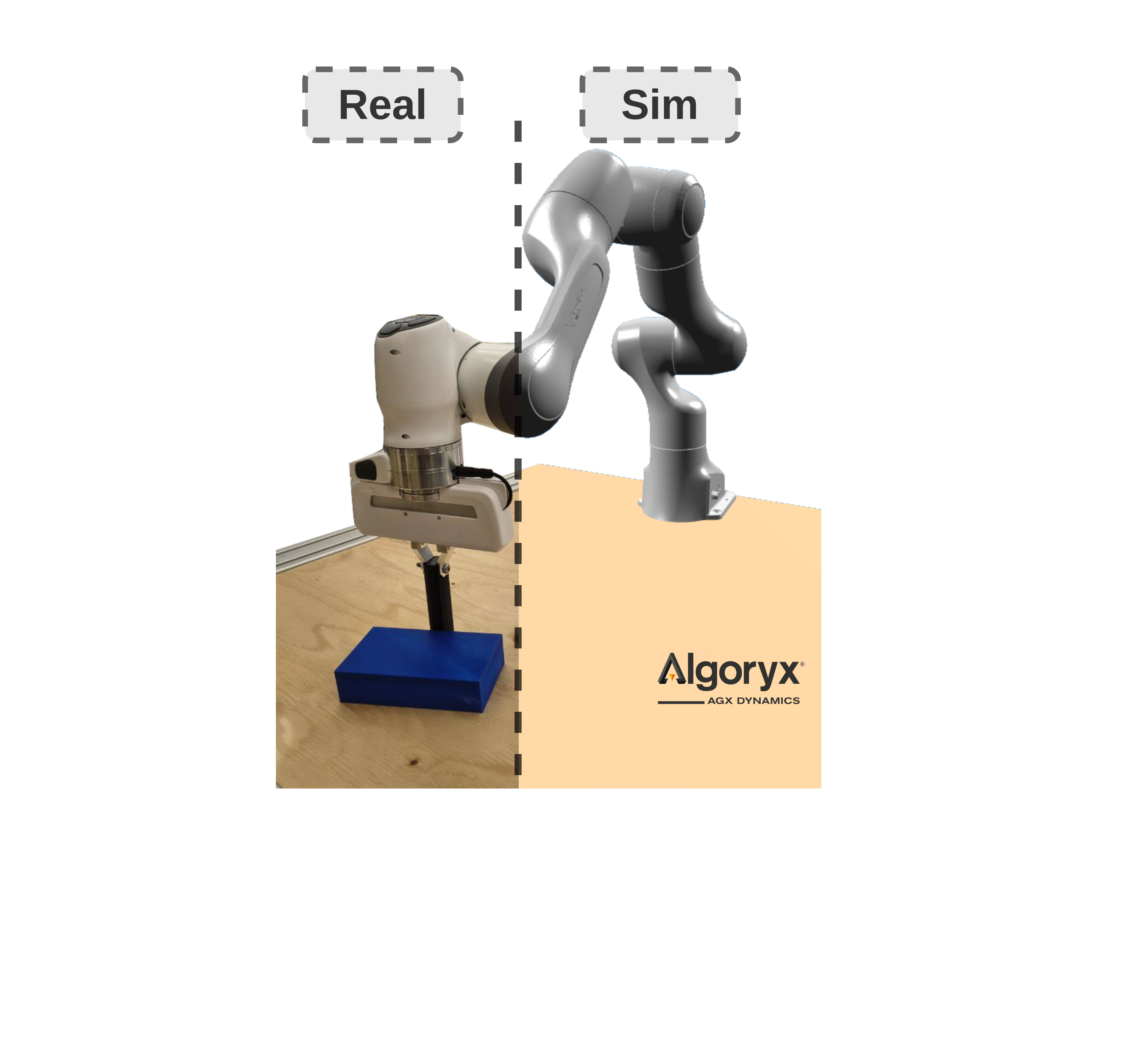} 
\caption{Setup employed for the experimental task evaluated in~\Cref{sec:eval_real} --- pushing a box to a desired location --- featuring the Franka Emika Panda robot and its digital twin in the AGX Dynamics simulation~\cite{agx}.}
\label{fig:robot_dt}
\vspace{-0.35cm}
\end{figure}

In parallel to this line of research, Zero-Shot Generalization (ZSG) has emerged as a key topic in RL and robotics~\cite{zsg_survey}.
ZSG aims to train policies that can generalize to novel environments, without requiring environment-specific training.
Within this area, some works have explored \textit{context}-aware policies --- policies that receive, alongside the state, an explicit input referred to as the \textit{context}, which captures structured information about the environment or task.
The assumption is that this context information, whether provided or inferred, enables the policy to adapt to varying conditions~\cite{contextualize_me}.
These approaches typically rely on an estimator trained to infer a context representation from experience, in a way analogous to OSI.
The key difference lies in the supervision strategy used during training, employing proxy tasks to enable context inference without direct supervision.

Despite sharing the objective of enabling policies to adapt to unseen dynamics, OSI and context-aware policies for ZSG have evolved in parallel, without direct comparison.
In this work, we aim to bridge these two lines of research and to investigate whether context-aware policies can enhance the sim-to-real transfer of robot control policies.
We focus on how the supervision strategy used to train the context estimator affects the policy's ability to generalize to unseen dynamics.
To this end, we conduct an empirical study of the main supervision strategies proposed in the literature, evaluated within a unified framework on both a canonical control benchmark and a real-robot task with a Franka Emika Panda, where the goal is to push a box to a desired location (\Cref{fig:robot_dt}).
In our benchmark scenarios, DR results in the poorest transfer performance under equivalent interaction samples.
In contrast, context-aware policies achieve better generalization, although no single method consistently outperforms the others in all settings.
This suggests that incorporating context information is beneficial, but that the optimal supervision strategy may be task-dependent.

This work makes two main contributions. 
First, we empirically demonstrate that incorporating context information and applying techniques from contextual RL results in improved sim-to-real transfer of manipulation policies.
Second, we analyze how different supervision strategies impact zero-shot generalization of context-conditioned policies, highlighting their task-dependent behavior and discussing practical implications for real-world deployment.


\section{Related Work}
\subsection{Sim-To-Real Transfer}
Domain Randomization is a common approach for transferring robot controllers trained with RL in simulation to reality~\cite{reality_gap_survey}.
The policy is trained across a diverse range of simulated environments by varying parameters that influence the environment dynamics, such as physical properties like mass and friction coefficients.
DR ensures that the policy is exposed to a broad spectrum of variations, enhancing its ability to perform robustly when deployed in the real world.

Several works have shown the effectiveness of DR in addressing the reality gap.
Matas et al.~\cite{sim_to_real_deformable_objects} train a controller in simulation for solving different deformable object manipulation tasks.
Experimental results indicate that randomizing extrinsic camera parameters aids sim-to-real transfer due to the controller's sensitivity to position changes, but excessive randomization can hinder transfer success.
Van Baar et al.~\cite{baar_str} show that DR-trained controllers require fewer fine-tuning steps for sim-to-real transfer in a robotic Marble Maze task.
OpenAI et al.~\cite{rubiks_cube} propose automatic domain randomization, where the environment parameters used during training are sampled from a changing distribution, rather than a fixed one.
\cite{reality_gap_survey} presents a comprehensive overview of approaches leveraging DR for sim-to-real transfer.
Common to most approaches is that the trained policies are not explicitly aware of the dynamics parameters randomized in simulation.
Instead, Yu et al.~\cite{online_system_identification} propose conditioning the control policy on these parameters, estimated by an online system identification model.
This model is trained in a supervised way using the ground truth parameters known in simulation and is designed to infer these parameters online at test time.
Despite promising results in simulation, the approach is not evaluated on a physical system, leaving its effectiveness for sim-to-real transfer unverified.
More broadly, the privileged information available in simulation (including ground-truth dynamics parameters) can be leveraged in different ways during training.
Beyond directly regressing these parameters, several works use privileged signals to supervise latent representations or to guide learning through teacher–student or asymmetric training schemes, where the student learns to reproduce teacher behavior or teacher features without requiring explicit parameter prediction~\cite{privileged_info1,privileged_info2}.
Collectively, these works demonstrate that leveraging ground-truth simulator signals as supervision --- whether through direct parameter regression or through representation learning --- is beneficial for sim-to-real transfer.
Finally, another line of research has proposed uncertainty-aware formulations of context-conditioned policies, where the latent context is represented not as a single point estimate but as a distribution or belief over possible dynamics~\cite{context_uncertainty}.
These approaches explicitly account for uncertainty in system identification and aim to produce policies that remain robust when the context cannot be reliably inferred from limited data.

In this work, we focus on point-estimate context inference.
We evaluate both the classical context-agnostic DR approach and a context-aware alternative that directly regresses the dynamic parameters through a learned context encoder, inspired by the system identification framework proposed in~\cite{online_system_identification}.

\subsection{Context-Aware Policies for Zero-Shot Generalization}
\label{context_aware_policies}
Recent approaches for ZSG of control policies build on the framework of Contextual RL (cRL)~\cite{contextual_rl, contextual_rl2}, which assumes variations in the environment can be represented by a \textit{context} and used to guide a generalizable agent in adapting its behavior accordingly.
Benjamins et al.~\cite{contextualize_me} show that optimal behavior in cRL requires context information.
This is validated by comparing the performance of a context-aware policy, which incorporates the known ground-truth context through simple state concatenation, with a context-agnostic policy.
Both policies are trained on various context-encoded versions of common RL environments, where dynamics parameters are randomized during training.
The results show that the context-conditioned policy often significantly outperforms the context-agnostic one, especially when the task is sensitive to changes in environment dynamics, underscoring the potential benefits of incorporating context.

In the more practically relevant case when a ground-truth context is not available, several approaches have been proposed to train a context estimator to generate a latent representation from recent transitions. 
This is similar to the system identification model introduced in~\cite{online_system_identification}, with the key difference being how the model (i.e., the context estimator) is supervised.
We identified two supervision strategies in the literature that can be applied to sim-to-real, where dynamics parameters are only available during training in simulation.
Evans et al.~\cite{iida} propose coupling the context estimator with a second model, referred to as the predictor, which utilizes the latent context to predict the next state of the environment.
In contrast, Ndir et al.~\cite{jcpl} propose training the context estimator based on the control policy loss, forcing the context to capture information relevant to the current policy.

In this work, we explore whether such latent context representations offer advantages for sim-to-real transfer and which supervision strategy yields the best results.
We train all policies in a unified framework and evaluate them both in simulation and on a physical setup, providing direct insight into their effectiveness for real-world deployment.


\section{Sim-to-real Transfer through Explicit Context Estimation}

We aim to investigate how conditioning RL policies on a representation of the environment's dynamics can improve sim-to-real transfer.
In particular, we explore the use of a \textit{context} vector $\mathbf{c} \in \mathbb{R}^c$, where $c \in \mathbb{N}^+$ denotes the context dimensionality, to capture unobservable but task-relevant environment properties, such as mass and friction, which vary between simulation and reality.
By leveraging this context information, the policy can adapt its behavior based on the specific dynamics of the deployment environment, rather than relying solely on robustness learned through domain randomization --- assuming simulation can approximate real-world dynamics through parameter tuning.

The approach consists of two main components: the \textit{control policy} $\pi$ and the \textit{context estimator} $\phi$.
The context estimator infers a context representation $\hat{\mathbf{c}} \in \mathbb{R}^{\hat{c}}$ from a set of $N$ transitions $(\mathbf{s}, \mathbf{a}, \mathbf{s}^{\prime})$, where $\hat{c} \in \mathbb{N}^+$, $\mathbf{s}$ is the state, $\mathbf{a}$ the control policy action, and $\mathbf{s}^{\prime}$ the next state:
\begin{equation}
    \label{eq:context_estimator}
    \phi \colon \{(\mathbf{s}_i, \mathbf{a}_i, \mathbf{s}^{\prime}_i)\}_{i=1}^{N} \mapsto \hat{\mathbf{c}}.
\end{equation}
%
%
Depending on the supervision strategy, $\hat{\mathbf{c}}$ may correspond to an estimate of the ground-truth context vector $\mathbf{c}$ or a latent representation learned through a proxy task.

The control policy $\pi$ receives both the observable state $\mathbf{s} \in \mathcal{S}$ and the inferred context representation $\hat{\mathbf{c}}$ as input, returning an action $\mathbf{a} \in \mathcal{A}$, corresponding to a robot command:
\begin{equation}
    \label{eq:control_policy}
    \pi \colon (\mathbf{s}, \hat{\mathbf{c}}) \mapsto \mathbf{a}.
\end{equation}

\subsection{Training and Evaluation}
\Cref{alg:training_loop} outlines the training procedure for a generic off-policy RL algorithm.
We jointly train the control policy $\pi$ and the context estimator $\phi$, following the approach proposed in~\cite{jcpl}.
A key advantage of this approach is that both $\pi$ and $\phi$ learn from data generated by the same policy $\pi$, eliminating the need for an auxiliary data collection policy as required in other methods.
Moreover, by using data from the same policy, we avoid distributional mismatch, ensuring that the context estimator is trained on state–action distributions that reflect those encountered during policy execution.

At the beginning of each episode, the agent receives an initial observation $\mathbf{s}_1$, while the episodic context $\mathbf{c}_e$, available in simulation, is retrieved.
At each time step $t$, we sample a set $\mathcal{T}_e$ of $N$ transitions sharing the ground-truth context $\mathbf{c}_e$ from the replay buffer.
Based on this set, the context estimator $\phi$ infers the context representation $\hat{\mathbf{c}}_e$.
The control policy then selects an action $\mathbf{a}_t$ based on the observable state \(\mathbf{s}_t\) and the inferred context $\hat{\mathbf{c}}_e$.
After action execution, the agent observes the next state $\mathbf{s}^{\prime}_t$ and receives a reward $r_t$, while the tuple $(\mathbf{s}_t, \mathbf{a}_t, r_t, \mathbf{s}^{\prime}_t, \mathbf{c}_e)$ is stored in the replay buffer.

To optimize the control policy and context estimator, we first sample a mini-batch $\mathcal{B}$ from the replay buffer.
For each transition in $\mathcal{B}$, we sample a set $\mathcal{T}_i$ of $N$ transitions sharing the context $\mathbf{c}_i$, following the same procedure used for action selection.
The context estimator $\phi$ then infers the context representation $\hat{\mathbf{c}}_i = \phi(\mathcal{T}_i)$ for each sampled transition.
Lastly, we compute the losses for the control policy and the context estimator and update them accordingly.
The context estimator loss is computed based on the selected supervision strategy (\cref{sec:sup_strategies}), while the control policy loss remains unchanged, depending only on the selected RL algorithm.

During evaluation, we follow the same procedure for selecting action $\mathbf{a}_t$, except that context estimation is performed using $N$ transitions sampled from the current episode.

\begingroup
\makeatletter
\renewcommand*{\@algocf@pre@ruled}{\vspace*{3pt}\hrule height .8pt depth0pt \kern2pt}
\makeatother

\begin{algorithm}[t!]
\small
\SetAlgoNoLine
Initialize control policy $\pi$, context estimator $\phi$, and replay buffer $\mathcal{R} = \emptyset$. \\
\For{$e=1, \dots, episodes$}{
    Receive observation $\mathbf{s}_1$ and context $\mathbf{c}_e$\\
    \For{$t=1, \dots, steps$ and $!done$}{
        Sample set $\mathcal{T}_e$ of $N$ transitions with context $\mathbf{c}_e$ from $\mathcal{R}$ \\

        Infer context representation $\hat{\mathbf{c}}_e = \phi(\mathcal{T}_e)$

        Select action $\mathbf{a}_t = \pi(\mathbf{s}_t, \hat{\mathbf{c}}_e)$ \\
        
        Execute $\mathbf{a}_t$, observe next state $\mathbf{s}^{\prime}_t$ and reward $r_t$ \\

        $\mathcal{R} = \mathcal{R} \cup (\mathbf{s}_t, \mathbf{a}_t, r_t, \mathbf{s}^{\prime}_t, \mathbf{c}_e)$ \\

        Sample mini-batch $\mathcal{B}$ from $\mathcal{R}$ \\
    
        \For{$(\mathbf{s}_i, \mathbf{a}_i, r_i, \mathbf{s}^{\prime}_i, \mathbf{c}_i) \in \mathcal{B}$}{
            Sample set $\mathcal{T}_i$ of $N$ transitions with context $\mathbf{c}_i$ from $\mathcal{R}$
    
            Infer context representation $\hat{\mathbf{c}}_i = \phi(\mathcal{T}_i)$ \\
        }

        Compute loss $\mathcal{L}_{\pi}$ and update $\pi$\\
        Compute loss $\mathcal{L}^{*}_{\phi}$ according to the selected supervision strategy and update $\phi$\\
    }
}
\label{alg:training_loop}
\caption{Training loop.}
\end{algorithm}

\subsection{Context Estimator Architectures}
Similarly to~\cite{iida}, we investigate three architectures for the context estimator.
The first employs a feed-forward neural network combined with average pooling (FF+AVG).
Each transition is processed through a shared estimator, and the resulting embeddings are averaged to produce a fixed-length representation.
This approach is simple and computationally efficient, though it treats all transitions equally, regardless of how informative they are.
The second architecture leverages a recurrent neural network, specifically an LSTM~\cite{lstm}, to aggregate context information.
Although the transitions are inherently unordered, we treat them as a sequence and extract the latent representation by applying a linear projection to the final hidden state of the LSTM.
This formulation allows the model to learn how to weigh and extract relevant information across different transitions, potentially improving its ability to capture complex environmental variations.
Finally, we consider a Transformer-based architecture~\cite{transformer}.
More specifically, since the transitions we consider for context estimation are unordered, we employ an order-invariant Transformer encoder, i.e., a self-attention architecture without positional encodings.
Each transition is first embedded into a fixed-dimensional token representation and processed jointly using multi-head self-attention.
A learnable aggregation token is prepended to the set of transition embeddings, and its output embedding is used as the final context representation.
This design allows the model to selectively attend to informative transitions while remaining permutation-invariant with respect to their ordering.

\subsection{Context Estimator Supervision Strategies}
\label{sec:sup_strategies}
We evaluate three strategies to supervise the context estimator training --- one based on regressing the ground truth context, and two on proxy tasks inspired by cRL research.

\textbf{Ground-Truth (GT) Supervision}.
Based on~\cite{online_system_identification}, the context estimator is trained to directly regress the ground-truth context available in simulation.
The loss is defined as the mean squared error (MSE) between the estimated and the ground-truth context vectors: 

\begin{equation}
    \mathcal{L}^{GT}_{\phi} = \mathbb{E}_{(\mathbf{s}, \mathbf{a}, \mathbf{s}^{\prime}, \mathbf{c}) \sim \mathcal{B}} \left[ \left\| \hat{\mathbf{c}}- \mathbf{c} \right\|^2 \right].
\end{equation}

\textbf{Proxy Task Forward Dynamics Prediction (FP)}.
Based on~\cite{iida}, the context estimator is trained end-to-end in conjunction with a prediction model on a forward prediction task.
The context estimator $\phi$ infers a context representation $\hat{\mathbf{c}}$ in a learned latent space (\Cref{eq:context_estimator}).
This latent vector is used to condition the prediction model $p_f$, which is implemented as a feed-forward neural network and infers the next state $\mathbf{s}^{\prime}$ from current state $\mathbf{s}$ and action $\mathbf{a}$:

\begin{equation}
    p_f \colon (\mathbf{s}, \mathbf{a}, \hat{\mathbf{c}}) \mapsto \mathbf{s}^{\prime}.
\end{equation}
$\phi$ and $p_f$ are updated jointly by minimizing the mean-squared error between the predicted and ground-truth next state, with gradient back-propagated through both models:

\begin{equation}
    \mathcal{L}^\mathit{FP}_{\phi} = \mathbb{E}_{(\mathbf{s}, \mathbf{a}, \mathbf{s}^{\prime}, \mathbf{c}) \sim \mathcal{B}} \left[ \left\| p_f\left(\mathbf{s}, \mathbf{a}, \hat{\mathbf{c}})\right) - \mathbf{s}^{\prime} \right\|^2 \right].
\end{equation}

\textbf{Proxy Task Policy Loss (PL)}.
Based on~\cite{jcpl}, the context estimator is trained by directly minimizing the policy loss, effectively using the policy's objective as a supervision signal.
Being the control policy explicitly conditioned on the context representation $\mathbf{\hat{c}}$ (\Cref{eq:control_policy}), $\mathcal{L}_{\pi}$ is differentiable with respect to the parameters of $\phi$.
Consequently, the gradients of the control policy loss can be backpropagated directly into the context estimator: 

\begin{equation}
    \mathcal{L}^\mathit{PL}_{\phi} = \mathcal{L}_{\pi},
\end{equation}
where $\mathcal{L}_{\pi}$ denotes the policy loss associated with the selected RL algorithm.
This implicit supervision drives $\phi$ to learn latent context representations that maximise the policy's expected return.


\section{Evaluation}
We evaluate the different supervision strategies on both sim-to-sim and sim-to-real transfer tasks.
In~\Cref{sec:eval_sim}, we consider a classic control task commonly used in RL, assessing the zero-shot generalization of policies across different simulated domains.
Although this setup does not involve a robot or actual sim-to-real transfer, it offers a low-cost and reproducible benchmark for comparison in controlled domain conditions.
In~\Cref{sec:eval_real}, we assess the sim-to-real transfer on a physical robot to evaluate the effectiveness of the strategies in real-world settings.
In all experiments, we employ the Soft Actor-Critic (SAC) algorithm~\cite{sac} for training the policies, in its implementation provided by \textit{Stable-Baselines3} \cite{stable_baselines3}.
We design the actor and critic networks with $2$ layers of $256$ neurons each, while the context estimator and the predictor with $2$ layers of $16$ neurons each.

\textbf{Baselines.}
We compare the supervision strategies described in~\Cref{sec:sup_strategies} with two baseline approaches.
The first, referred to as \textit{Oracle}, augments the policy’s input by directly appending the ground truth context to the observable state.
While this approach is not applicable to sim-to-real transfer, since the ground-truth context is not available in real-world settings, it serves as a reference profile to assess the performance of the other policies in simulation.
In contrast, the second baseline, referred to as \textit{Agnostic}, restricts the policy’s input solely to the observable state, reflecting the conventional domain randomization approach that does not incorporate explicit context information.

\textbf{Evaluation.}
To assess the generalization capability of the trained policies, we generate three distinct context sets: training, validation, and test.
These sets are obtained using Latin hypercube sampling with different seeds to ensure coverage and variability within each context space.
All policies are trained on the same training set by iteratively cycling through the available contexts, with a different context assigned to each training episode in a round-robin fashion.
In contrast to standard RL tasks, which typically allow periodic evaluation on a limited number of episodes in a fixed environment, our framework requires performance assessment across a broad spectrum of context values.
Consequently, frequent policy evaluation during training on such an extensive validation set is computationally infeasible.
To address this challenge, we adopt a sparser evaluation strategy.
After the training, we select a limited number of checkpoints, evenly spaced over a predefined interval of training steps.
Among these, the best-performing checkpoint on the validation set is then evaluated on the test set, and we report the corresponding results in the tables.
We provide the details regarding the training, validation, and test sets, as well as the checkpoint selection interval, in each experiment's description.

\subsection{Classic Control Task}
\label{sec:eval_sim}
\textbf{Description.}
We use the CARL library~\cite{contextualize_me}, which provides contextual extensions to standard RL environments by enabling systematic variation of domain parameters such as mass, friction, and damping.
For our experiments, we select the \textit{Pendulum} environment and consider gravity magnitude $g$, pendulum length $l$, and mass $m$ as context parameters.

We conduct three main experiments.
The first experiment investigates how the context dimensionality affects generalization.
To this end, we evaluate all possible combinations of the three context parameters, ranging from $1D$, where each parameter is considered in isolation, to $3D$, where all three parameters vary jointly.
This results in a total of seven context space combinations.
In line with~\cite{contextualize_me}, we define the bounds of each context feature as $0.1$ to $2$ times its standard value.
To enable a fair comparison across different context dimensions, we maintain a consistent sampling density by increasing the number of samples exponentially with the number of dimensions.
Specifically, we use $7$, $49$, and $343$ samples for the $1D$, $2D$, and $3D$ spaces, respectively.
We run $10^5$ training steps in the case of the $1D$ and $2D$ context spaces, and $2 \times 10^5$ for the $3D$ one.
For all the supervision strategies but \textit{GT}, we fix the dimension of the context representation to the number of context dimensions plus one.
This choice follows common practice in related work, where a slightly over-parameterized latent space is used to provide additional representational capacity, which can improve policy performance and generalization.
We select $50$ checkpoints over the final $25\%$ of training steps for evaluation, and we run $3$ episodes per context for both evaluation and test.
In~\Cref{tab:pendulum1}, we report the average and the best test performance in terms of reward return, computed over $10$ replicates obtained by varying the seeds for both policy initialization and environment setup.

The second experiment investigates the effect of varying the dimensions of the context representation.
We fix the context dimensionality to the $3D$ case and train the \textit{FP} and \textit{PL} policies using context representation dimensions ranging from $2$ to $6$.
As in the first experiment, we generate three sets for training, validation, and testing, performing validation and testing as previously described.
\Cref{tab:pendulum2} shows the average and the best results on the test set over $10$ replicates.

Finally, the third experiment investigates the effect of varying the number of input transitions for context estimation.
We fix the context dimensionality to the $3D$ case and train all context estimation-based policies using different numbers of input transitions, specifically $10$, $15$, $20$, $25$, and $30$.
Training, validation, and testing are performed following the same protocol as in the previous experiments.
\Cref{tab:pendulum3} reports both the average and the best test performance over $10$ independent runs.

\textbf{Analysis.}
Across all evaluated context configurations reported in \Cref{tab:pendulum1}, conditioning on explicit contextual information --- either via a ground-truth oracle or through a learned estimator --- results in markedly improved performance compared to the \textit{Agnostic} baseline.
\textit{Oracle} establishes an empirical upper bound on performance, exhibiting very low variance across runs.
In no case does a learned estimator exceed \textit{Oracle}'s performance.
While the best-performing runs across methods can occasionally approach \textit{Oracle}’s return, none consistently achieve superior performance.

Among the learned supervision strategies, both \textit{GT} and \textit{PL} exhibit comparable and consistently superior performance.
We conduct a Welch’s t-test to evaluate the statistical significance of the observed performance differences between \textit{GT} and \textit{PL}, comparing corresponding architectures (LSTM and Transformers), due to its robustness to unequal variances across samples.
For the LSTM-based estimators, the resulting p-values exceed the conventional significance threshold ($p > 0.05$) across all context settings, indicating that the differences are not statistically significant.
For the Transformer-based estimators, statistically significant differences emerge in some higher-dimensional context settings --- specifically for the (\textit{g}, \textit{m}), (\textit{l}, \textit{m}) and (\textit{g}, \textit{l}, \textit{m}) combinations --- while remaining non-significant in the other cases.
Overall, these results suggest that weak supervision via policy loss can be as effective as fully supervised regression, but that the relative behavior of the two strategies may depend on the choice of context estimator architecture and the dimensionality of the context space.

\begin{landscape}
\begin{table}[p]
    \vspace{4pt}
    \centering
    \scriptsize
    \renewcommand\arraystretch{1.3}
    \caption{Test reward on Pendulum with varying context. Numbers encode mean $\pm$ std (best) across seeds.}
    \begin{tabular}{@{\hspace{5pt}}c@{\hspace{5pt}}c@{\hspace{5pt}}|@{\hspace{5pt}}c@{\hspace{8pt}}c@{\hspace{8pt}}c@{\hspace{5pt}}|@{\hspace{5pt}}c@{\hspace{8pt}}c@{\hspace{8pt}}c@{\hspace{5pt}}|@{\hspace{5pt}}c@{\hspace{5pt}}}
    \multicolumn{2}{c}{\textbf{Policy}} & \multicolumn{3}{c}{\textbf{1D Context}} & \multicolumn{3}{c}{\textbf{2D Context}} & \multicolumn{1}{c}{\textbf{3D Context}}\\ [0.5ex]

     & estimator & g & l & m & g, l & g, m & l, m & g, l, m \\
     
     \toprule
     
     \textit{\textbf{Oracle}} & \textit{-} & \textit{-280 $\pm$ 1 (-279)} & \textit{-160 $\pm$ 2 (-154)} & \textit{-148 $\pm$ 3 (-144)} & \textit{-354 $\pm$ 13 (-333)} & \textit{-358 $\pm$ 5 (-346)} & \textit{-240 $\pm$ 7 (-229)} & \textit{-384 $\pm$ 7 (-376)} \\
     \midrule
     \textbf{Agnostic} & - & -355 $\pm$ 36 (-307) & -348 $\pm$ 35 (-274) & -175 $\pm$ 7 (-163) & -633 $\pm$ 25 (-595) & -490 $\pm$ 18 (-474) & -530 $\pm$ 42 (-467) & -575 $\pm$ 24 (-543) \\
     \midrule
     \multirow{2}{*}{\textbf{GT}} & FF+AVG & -313 $\pm$ 16 (-289) & -199 $\pm$ 23 (-171) & -374 $\pm$ 179 (-157) & -488 $\pm$ 37 (-447) & -463 $\pm$ 39 (-407) & -458 $\pm$ 78 (-358) & -522 $\pm$ 36 (-470)\\
      & LSTM & -310 $\pm$ 11 (-300) & -191 $\pm$ 9 (-175) & -173 $\pm$ 17 (-151) & -464 $\pm$ 66 (-378) & -405 $\pm$ 40 (-362) & -298 $\pm$ 41 (-242) & -448 $\pm$ 31 (-389)\\
      & TF & \textbf{-305 $\pm$ 10} (-292) & -212 $\pm$ 28 (-169) & \textbf{-163 $\pm$ 6} (-152) & \textbf{-407 $\pm$ 32} (-373) & -384 $\pm$ 18 (-356) & -283 $\pm$ 30 (-240) & -420 $\pm$ 15 (-394)\\
      \midrule
     \multirow{3}{*}{\textbf{FP}} & FF+AVG & -362 $\pm$ 129 (-296) & -396 $\pm$ 152 (-209) & -176 $\pm$ 12 (-156) & -611 $\pm$ 156 (-417) & -491 $\pm$ 71 (-410) & -534 $\pm$ 152 (-412) & -596 $\pm$ 79 (-509)\\
      & LSTM & -326 $\pm$ 22 (-298) & -216 $\pm$ 14 (-188) & -199 $\pm$ 28 (-167) & -496 $\pm$ 42 (-428) & -400 $\pm$ 33 (-371) & -409 $\pm$ 80 (-314) & -465 $\pm$ 31 (-401)\\
      & TF & -316 $\pm$ 24 (-290) & -240 $\pm$ 59 (-177) & -175 $\pm$ 19 (-151) & -496 $\pm$ 24 (-454) & -391 $\pm$ 28 (-360) & -356 $\pm$ 32 (-297) & -415 $\pm$ 22 (-377)\\
      \midrule
      \multirow{3}{*}{\textbf{PL}} & FF+AVG & -430 $\pm$ 137 (-300) & -271 $\pm$ 40 (-199) & -361 $\pm$ 167 (-151) & -522 $\pm$ 32 (-448) & -450 $\pm$ 52 (-379) & -465 $\pm$ 115 (-281) & -511 $\pm$ 61 (-410) \\
      & LSTM & -308 $\pm$ 16 (-298) & \textbf{-181 $\pm$ 24} (-163) & -176 $\pm$ 46 (-150) & -461 $\pm$ 71 (-376) & -390 $\pm$ 21 (-360) & -291 $\pm$ 39 (-235) & -441 $\pm$ 81 (-379) \\
      & TF & -318 $\pm$ 17 (-295) & -221 $\pm$ 23 (-183) & -165 $\pm$ 18 (-151) & -418 $\pm$ 57 (-348) & \textbf{-374 $\pm$ 16} (-353) & -2\textbf{58 $\pm$ 16} (-235) & \textbf{-382 $\pm$ 15} (-365)\\
      
      \bottomrule
        
    \end{tabular}
    \label{tab:pendulum1}
    \vspace{-0.2cm}
\end{table}

\begin{table}[p]
    \centering
    \scriptsize
    \renewcommand\arraystretch{1.2}
    \caption{Test reward on Pendulum with varying latent context dimension. Numbers encode mean $\pm$ std (best) across seeds.}
    \begin{tabular}{cc|ccccc}
    \multicolumn{2}{c}{\textbf{Policy}} & \multicolumn{5}{c}{\textbf{Latent Context Dimensions}}\\ [0.5ex]

     &  estimator & 2 & 3 & 4 & 5 & 6\\
     
     \toprule

     \multirow{3}{*}{\textbf{FP}} & FF+AVG & -608 $\pm$ 68 (-490) & -628 $\pm$ 62 (-496) & -596 $\pm$ 79 (-509) & -589 $\pm$ 69 (-510) & -563 $\pm$ 74 (-471)\\
      & LSTM & -496 $\pm$ 25 (-455) & -522 $\pm$ 71 (-456) & -465 $\pm$ 31 (-401) & -449 $\pm$ 29 (-407) & -460 $\pm$ 47 (-378)\\
      & TF & -482 $\pm$ 41 (-414) & -454 $\pm$ 20 (-423) & -415 $\pm$ 22 (-377) & -436 $\pm$ 31 (-392) & -425 $\pm$ 31 (-383)\\
      \midrule
     \multirow{3}{*}{\textbf{PL}} & FF+AVG & -495 $\pm$ 42 (-432) & -497 $\pm$ 43 (-457) & -511 $\pm$ 61 (-410) & -467 $\pm$ 47 (-398) & -493 $\pm$ 43 (-426)\\
      & LSTM & -466 $\pm$ 39 (-411) & -499 $\pm$ 93 (-425) & -441 $\pm$ 81 (-379) & -439 $\pm$ 37 (-380) & -438 $\pm$ 35 (-392)\\
      & TF & \textbf{-432 $\pm$ 29} (-384) & \textbf{-436 $\pm$ 37} (-416) & \textbf{-382 $\pm$ 15} (-365) & \textbf{-382 $\pm$ 14} (-363) & \textbf{-398 $\pm$ 40} (-360)\\

      \bottomrule
        
    \end{tabular}
    \label{tab:pendulum2}
    \vspace{-0.35cm}
\end{table}

\begin{table}[p]
    \centering
    \scriptsize
    \renewcommand\arraystretch{1.2}
    \caption{Test reward on Pendulum with varying number of input transitions for context estimation. Numbers encode mean $\pm$ std (best) across seeds.}
    \begin{tabular}{cc|ccccc}
    \multicolumn{2}{c}{\textbf{Policy}} & \multicolumn{5}{c}{\textbf{Number of Input Transitions}}\\ [0.5ex]

     &  estimator & 10 & 15 & 20 & 25 & 30\\
     
     \toprule

     \multirow{3}{*}{\textbf{GT}} & FF+AVG & -505 $\pm$ 53 (-420) & -482 $\pm$ 48 (-420) & -522 $\pm$ 36 (-470) & -551 $\pm$ 66 (-427) & -539 $\pm$ 56 (-447)\\
      & LSTM & -433 $\pm$ 19 (-404) & -446 $\pm$ 25 (-406) & -448 $\pm$ 31 (-389) & -459 $\pm$ 44 (-395) & -447 $\pm$ 40 (-391)\\
      & TF & -391 $\pm$ 15 (-356) & -406 $\pm$ 15 (-384) & -420 $\pm$ 15 (-394) & -415 $\pm$ 15 (-396) & -432 $\pm$ 30 (-395)\\
      \midrule

      \multirow{3}{*}{\textbf{FP}} & FF+AVG & -513 $\pm$ 48 (-429) & -572 $\pm$ 69 (-470) & -596 $\pm$ 79 (-509) & -575 $\pm$ 52 (-532) & -596 $\pm$ 45 (-516)\\
      & LSTM & -462 $\pm$ 30 (-418) & -467 $\pm$ 36 (-422) & -465 $\pm$ 31 (-401) & -466 $\pm$ 35 (-423) & -477 $\pm$ 47 (-432)\\
      & TF & -405 $\pm$ 18 (-364) & -413 $\pm$ 18 (-378) & -415 $\pm$ 22 (-377) & -409 $\pm$ 15 (-374) & -423 $\pm$ 18 (-397)\\
      \midrule

      \multirow{3}{*}{\textbf{PL}} & FF+AVG & -472 $\pm$ 30 (-407) & -483 $\pm$ 31 (-421) & -511 $\pm$ 61 (-410) & -491 $\pm$ 52 (-435) & -510 $\pm$ 37 (-468)\\
      & LSTM & -476 $\pm$ 55 (-399) & -405 $\pm$ 12 (-380) & -441 $\pm$ 81 (-379) & -468 $\pm$ 70 (-386) & -432 $\pm$ 31 (-392)\\
      & TF & \textbf{-380 $\pm$ 22} (-348) & \textbf{-382 $\pm$ 27 (-348)} & \textbf{-382 $\pm$ 15} (-365) & \textbf{-388 $\pm$ 26} (-350) & \textbf{-411 $\pm$ 20} (-383)\\

      \bottomrule
        
    \end{tabular}
    \label{tab:pendulum3}
    \vspace{-0.35cm}
\end{table}
\end{landscape}

Analysis of performance across increasing context dimensionality reveals a degradation in both return and stability for all methods, with broader confidence intervals and growing divergence from \textit{Oracle}.
This is particularly pronounced in the $3D$ case, underscoring the difficulty of accurately inferring context embeddings in high-dimensional settings.

Ablation results provided in~\Cref{tab:pendulum2} further examine the sensitivity of estimator performance to the dimensionality of the latent embedding space.
For both \textit{FP} and \textit{PL}, performance improves as the latent dimension increases from $2$ to $5$, after which it either saturates or exhibits minor deterioration.
This observation aligns with prevailing heuristics that advocate for latent spaces whose dimensionality slightly exceeds that of the ground-truth context.
On the other side, ablation results provided in~\Cref{tab:pendulum3} confirm that the context window size has only a minor effect on performance for most configurations, with smaller windows performing slightly better in the majority of cases.
The degradation from 10 to 30 input transitions is most pronounced for FF+AVG estimators, where the effect is consistent across all three supervision strategies.
Notably, \textit{PL} with TF-based context estimator achieves nearly identical performance for windows of $10$, $15$, $20$ and $25$ transitions before degrading at 30.
We attribute the general trend towards better performance with smaller windows to the evaluation protocol: a larger context window requires the policy to act for a greater number of steps on incomplete context information, until a sufficient number of transitions has been accumulated to match the training regime, which in turn means that critical decisions early in the episode may need to be taken based on faulty context estimates.

Finally, across methods and context settings, architectures that can selectively aggregate information across transitions --- namely the LSTM and the Transformer --- consistently achieve higher and more reliable performance than the feed-forward estimator with average pooling (FF+AVG).
The Transformer matches or exceeds the LSTM in several settings, indicating that attention-based aggregation can be at least as effective as recurrence for context estimation.
In contrast, while FF+AVG can occasionally reach comparable best-case performance in low-dimensional contexts, such cases are uncommon and exhibit substantially higher variance, suggesting strong sensitivity to initialization and training dynamics.
Overall, both LSTM and Transformer estimators yield more consistent results, with the Transformer emerging as the most robust and consistently high-performing choice among the evaluated architectures.

\subsection{Pushing Task}
\label{sec:eval_real}

\begin{figure}[t!]
\centering
\includegraphics[width=0.65\linewidth,trim={0 0.5cm 1cm 0},clip]{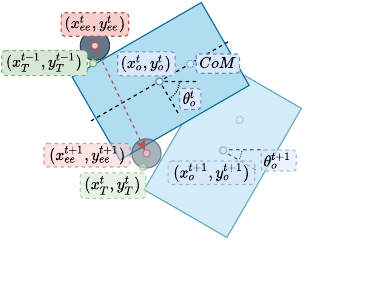} 
\vspace{-0.2cm}
\caption{Illustration of the planar pushing task, showing the object and end-effector positions at two consecutive time steps. Solid blue dots represent the object positions, while dashed blue dots represent its center of mass. Red circles represent the position of the cylindrical tool mounted on the robot hand, which is used to push the object. Green circles represent the target end-effector positions issued to the Cartesian Impedance Controller to command the robot end-effector.}
\label{fig:pushing_ill}
\vspace{-0.35cm}
\end{figure}

\textbf{Description.}
We evaluate sim-to-real transfer on a planar pushing task, where a robot arm uses its end-effector to push a box from an initial to a desired end position (\Cref{fig:pushing_ill}).
This task has been extensively studied in the context of sim-to-real transfer, due to its sensitivity to latent physical properties such as mass and friction~\cite{pushing,pushing_bergmann}.
We model the problem in the commercial physics engine AGX Dynamics~\cite{agx}, using a high-fidelity digital twin of the Franka Emika Panda robot employed for the real-world experiments (\Cref{fig:robot_dt}).
We use a Cartesian Impedance Controller~\cite{impedance_contr} to learn a policy that directly operates in the Cartesian space, enabling compliant interaction with the environment through force-aware motion control.
We define observation and action at time-step $t$ as follows:
\begin{equation}
    \label{eq:pushing_observation}
	\mathbf{s}^{t} = [x^{t}_{ee}, y^{t}_{ee}, x^{t}_o, y^{t}_o, \theta^{t}_o, \Delta x^{t-1}_T, \Delta y^{t-1}_T],
\end{equation}
\begin{equation}
    \label{eq:pushing_action}
    \mathbf{a}^{t} = [\Delta x^{t}_T, \Delta y^{t}_T],
\end{equation}
where:
\begin{itemize}
    \item $x^t_{ee}$ and $y^t_{ee}$ denote the end-effector 2D position;
    \item $x^t_o, y^t_o, \theta_o^t$ denote the object 2D position and planar orientation;
    \item $\Delta x^t_T, \Delta y^t_T$ denote the target control action, defined as the displacement relative to the previous action, computed as $\Delta v^t = v^t - v^{t-1}$ for $v \in \{x_T, y_T\}$.
\end{itemize}

We employ the following reward function, which smoothly penalizes the distance to the goal:
\begin{equation}
    \label{eq:pushing_reward}
    r = -\log\left(1 + \frac{d}{\delta}\right) + \mathbb{I}_{\text{fail}} \cdot r_{\text{fail}}
\end{equation}
where:
\begin{itemize}
    \item $d$ is the Euclidean distance between the object and the goal position;
    \item $delta$ is a normalization constant;
    \item $r_{\text{fail}}$ is a fixed penalty applied upon failure (i.e., kinematic infeasibility due to unreachable positions).
\end{itemize}
The task is considered successful when the object is within a fixed threshold of $3\,\mathrm{cm}$ from the target location.
We truncate episodes after $250$ steps.

\begin{table}[t!]
    \vspace{5pt}
    \centering
    \small
    \renewcommand\arraystretch{1.2}
    \caption{Randomized parameters for the pushing task}
    \begin{tabular}{ll}
    \textbf{Parameter} & \textbf{Sampling Distribution}\\ [0.5ex]     
    \toprule
    Box Mass                    & $\mathcal{U}([0.1, 1])\,\mathrm{kg}$\\
    Box-Tool Friction           & $\mathcal{U}([0.1, 0.5])$\\
    Box-Table Friction          & $\mathcal{U}([0.2, 0.8])$\\
    Box Center of Mass     & $\mathcal{U}([-0.04, 0.04])\,\mathrm{m}$ (rel. to centroid)\\
    \midrule
    Starting Robot Position ($x$) & $\mathcal{U}([0.45, 0.55])\,\mathrm{m}$ \\
    Starting Robot Position ($y$) & $\mathcal{U}([0.25, 0.35])\,\mathrm{m}$ \\

    Starting Box Position ($x$) & $\mathcal{U}([-0.03, 0.03])\,\mathrm{m}$ (rel. to robot) \\
    Starting Box Position ($y$) & $\mathcal{U}([-0.1025, -0.0675])\,\mathrm{m}$ (rel. to robot) \\
    Starting Box Orientation & $\mathcal{U}([-0.5236, 0.5236])\,\mathrm{rad}$ \\
    \midrule
    Action Duration & $\mathcal{U}(\{\, 0.04 + 0.005\, i \,\}_{i=0}^4\})\,\mathrm{s}$ \\
    Box Position Noise          & $\mathcal{N}(0, 0.003^2)\,\mathrm{m}$\\
    Box Orientation Noise       & $\mathcal{N}(0, 0.05^2)\,\mathrm{rad}$\\
    
    \bottomrule
        
    \end{tabular}
    \label{tab:rand_pushing}
    \vspace{-0.35cm}
\end{table}

We conduct two experiments using a box of fixed dimensions $17\,\mathrm{cm} \times 10.5\,\mathrm{cm}$ and a cylindrical tool, measuring $10.8\,\mathrm{cm}$ in length and $3\,\mathrm{cm}$ in diameter, which is mounted on the robot hand and used to push the object.
In the first experiment, the context includes the mass of the box, the friction coefficient between the box and the tool, and the friction coefficient between the box and the table.
In the second experiment, we additionally include the box's center of mass, varying its position along the longer axis of the box.
To enhance sim-to-real transfer, we randomize the starting box pose and end-effector position within a pre-defined workspace region, and we perturb the box pose with Gaussian noise to simulate the inaccuracies introduced by the tracking system used in the real-world setup.
Additionally, we randomize action duration by discretely sampling the number of simulation steps, while keeping a fixed simulation step of $0.005\,\mathrm{s}$ to ensure consistent and stable physics solver performance. 
\Cref{tab:rand_pushing} shows the full list of randomized parameters and corresponding sampling distributions.

We train and evaluate all policies in simulation, and test them in simulation and on the physical robot.
Given the increased complexity of this task compared to the pendulum, we adopt a sampling strategy that mirrors the common $80/10/10$ ratio for training, validation, and test.
Specifically, we sample $400$ context values for training, $50$ for validation, and $50$ for testing in simulation from the defined context space.
We train each policy for $10^6$ steps and evaluate $200$ checkpoints over the second half of training steps, running $2$ episodes per context value for both evaluation and test in simulation.
Based on findings in~\Cref{sec:eval_sim}, we adopt an LSTM as the context estimator architecture and we fix the dimension of the context representation to the number of context dimension plus one --- resulting in 4 and 5 dimensions for the variation without and with the center of mass, respectively.
On the real robot, we evaluate the policies using $12$ context configurations obtained by combining $3$ surface materials with $4$ box variants.
\begin{figure}[t!]
\vspace{5pt}
\centering
\includegraphics[width=0.8\linewidth,trim={0 3cm 0cm 0.5cm},clip]{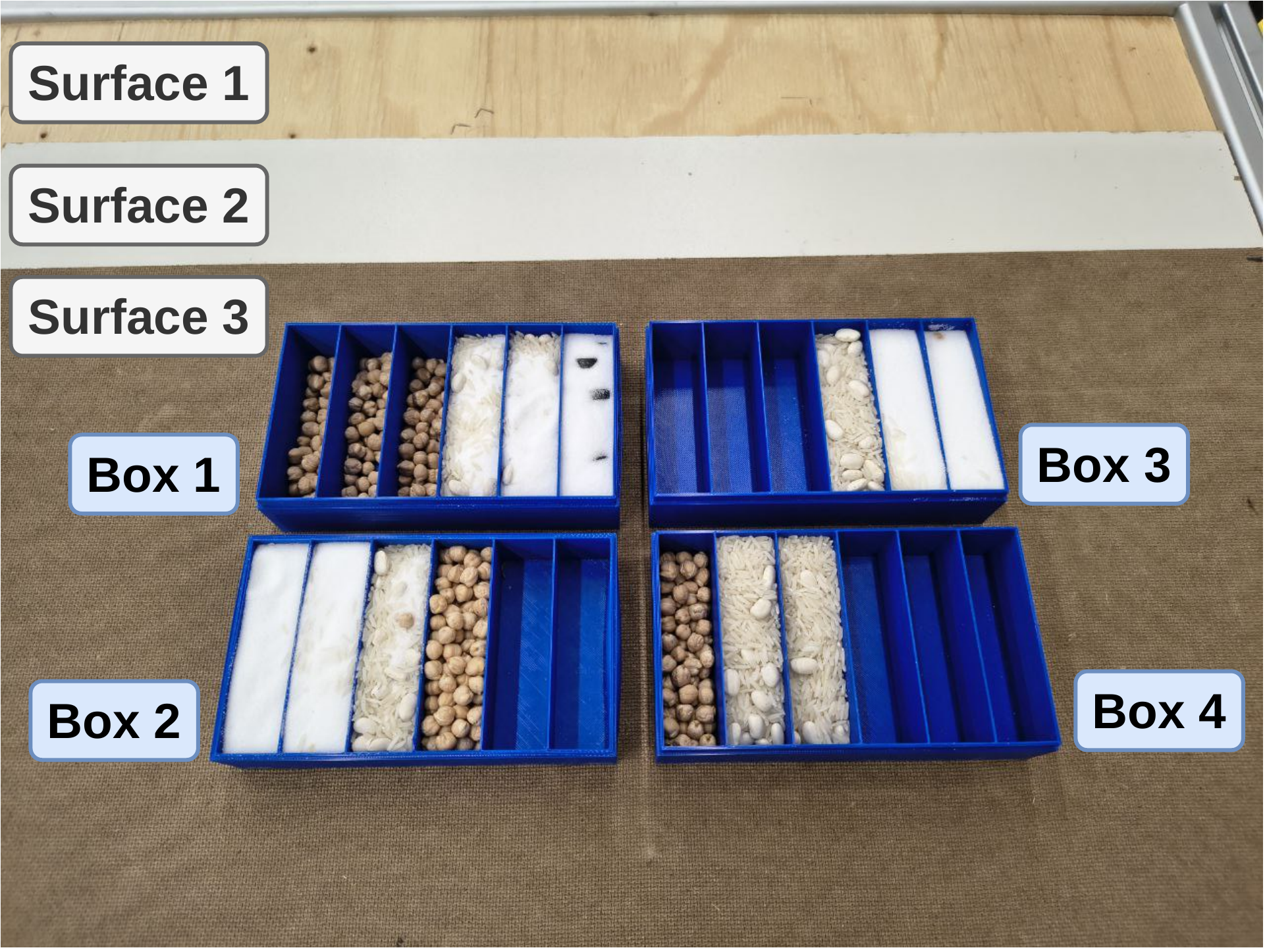} 
\caption{Context configurations used for real-world evaluation of the pushing task with center of mass variation, obtained by combining 3 surface materials 4 box variants. The surface materials differ in their friction coefficients, while the boxes vary in both mass and mass distribution, achieved by redistributing the filling material using internal separators.}
\label{fig:pushing_contexts_com}
\vspace{-0.35cm}
\end{figure}
For the setting without center of mass variation, the $4$ box variants differ in mass only, achieved by filling the box with different materials, ranging from $283\,\mathrm{g}$ to $824\,\mathrm{g}$.
For the other setting, the $4$ variants differ in both mass and mass distribution, achieved by using internal separators to distribute the filling material non-uniformly, with mass ranging from $276\,\mathrm{g}$ to $803\,\mathrm{g}$ (\Cref{fig:pushing_contexts_com}).
For each of the $12$ context configurations, we run $5$ episodes, resulting in a total of $60$ episodes per test.
We employ FoundationPose~\cite{foundationposewen2024} to track the box pose during real-world experiments.
\Cref{tab:robot_exps} shows the average and best results on the test sets over $3$ replicates.

\begin{landscape}
\begin{table}[p]
    \vspace{4pt}
    \centering
    \scriptsize
    \caption{Test reward on pushing. Numbers show mean $\pm$ std (best) across seeds.}
    \renewcommand\arraystretch{1.1}
    \begin{tabular}{c|cccc|cccc}
        \multicolumn{1}{c}{\textbf{Policy}} & \multicolumn{4}{c}{\textbf{Without Center of Mass}} & \multicolumn{4}{c}{\textbf{With Center of Mass}}\\ [0.5ex]
        & \multicolumn{2}{c}{Sim} & \multicolumn{2}{c}{Real} & \multicolumn{2}{c}{Sim} & \multicolumn{2}{c}{Real} \\
        & Reward & Success Rate & Reward & Success Rate & Reward & Success Rate & Reward & Success Rate \\
        \toprule

        \multirow{2}{*}{\textbf{\textit{Oracle}}} & \textit{-286 $\pm$ 3} & \textit{0.97 $\pm$ 0.03} & \textit{-} & \textit{-} & \textit{-299 $\pm$ 12} & \textit{0.95 $\pm$ 0.02} & \textit{-} & \textit{-} \\
        
        & \textit{(-283)} & \textit{(1.00)} & \textit{-} & \textit{-} & \textit{(-282)} & \textit{(0.97)} & \textit{-} & \textit{-} \\
        \midrule
        \multirow{2}{*}{\textbf{Agnostic}} & -287 $\pm$ 8 & 0.93 $\pm$ 0.14 & -359 $\pm$ 2 & 0.69 $\pm$ 0.03 & -381 $\pm$ 22 & 0.68 $\pm$ 0.07 & -541 $\pm$ 32 & 0.32 $\pm$ 0.08 \\
        & (-276) & (0.94) & (-357) & (0.72) & (-362) & (0.76) & (-502) & (0.40) \\
        \midrule
        \multirow{2}{*}{\textbf{GT}} & -273 $\pm$ 10 & 0.97 $\pm$ 0.02 & -340 $\pm$ 24 & 0.71 $\pm$ 0.13 & -403 $\pm$ 32 & 0.67 $\pm$ 0.11 & -570 $\pm$ 25 & 0.33 $\pm$ 0.05 \\
        & (-266) & (0.99) & (-323) & (0.82) & (-378) & (0.82) & (-536) & (0.40) \\
        \midrule
        \multirow{2}{*}{\textbf{FP}} & \textbf{-269 $\pm$ 13} & \textbf{0.99 $\pm$ 0.05} & \textbf{-315 $\pm$ 13} & \textbf{0.78 $\pm$ 0.04} & \textbf{-351 $\pm$ 24} & \textbf{0.78 $\pm$ 0.06} & \textbf{-496 $\pm$ 34} & \textbf{0.48 $\pm$ 0.07} \\
        & (-256) & (1.00) & (-297) & (0.82) & (-326) & (0.86) & (-448) & (0.58) \\
        \midrule
        \multirow{2}{*}{\textbf{PL}} & -277 $\pm$ 16 & 0.97 $\pm$ 0.01 & -322 $\pm$ 17 & 0.78 $\pm$ 0.10 & -475 $\pm$ 180 & 0.59 $\pm$ 0.39 & -592 $\pm$ 89  & 0.31 $\pm$ 0.24 \\
        & (-257) & (0.99) & (-302) & (0.92) & (-340) & (0.88) & (-485) & (0.63) \\
        \bottomrule
    \end{tabular}%
    \label{tab:robot_exps}
    \vspace{-0.2cm}
\end{table}
\end{landscape}

\textbf{Analysis.}
The transfer from simulation to the real robot leads to a performance degradation across all evaluated policies, highlighting the inherent challenge of zero-shot sim-to-real transfer.
Nonetheless, policies trained with access to contextual information consistently outperform the \textit{Agnostic} baseline, regardless of the context supervision strategy.
Importantly, the relative performance ranking observed in simulation is reflected in the real-world evaluations, indicating consistency between simulated and physical deployments.

In the scenario without center-of-mass variation, all context-aware policies outperform the \textit{Agnostic} baseline.
Among these, \textit{FP} yields the highest average return and exhibits minimal variability across seeds, indicating consistent performance.
\textit{PL} achieves slightly lower return metrics than \textit{FP}, both on average and in the best-performing seed, but attains the highest success rate for its top-performing policy.
As illustrated in \Cref{fig:box_trajectories}, this discrepancy arises because \textit{FP} completes successful episodes in fewer steps and ends unsuccessful ones closer to the goal, resulting in higher overall returns despite a lower success rate.
\Cref{fig:pushing_fp} shows representative trajectories on the real robot generated by the \textit{FP} policy for two different center-of-mass configurations.
\textit{GT} performs the worst among the context-aware methods both in average and best return metrics, while achieving a similar success rate when compared to \textit{FP}.

In the task variant including the center of mass variation, the performance gap between the \textit{Agnostic} baseline and context-aware methods becomes more pronounced.
Notably, \textit{GT} performs worse than the \textit{Agnostic} baseline, making this the only scenario where a context-conditioned policy underperforms the context-agnostic counterpart in our experiments.
\textit{FP} maintains consistent performance across seeds, exhibiting limited degradation despite the increased task complexity introduced by the center-of-mass variation. It achieves the highest return, both in terms of average performance and best-performing seed, mirroring the trend observed in the simpler task variant. However, as in the previous scenario, it does not attain the highest success rate.
\textit{PL}, on the other hand, reaches competitive performance in one of the seeds but shows substantially higher variance compared to the other methods.
This variability is attributed to one of the three training runs failing to converge to a competent policy, even in simulation. Despite this, \textit{PL} yields the highest success rate among all evaluated methods.

\begin{figure}[t!]
\centering
\includegraphics[width=0.99\linewidth,trim={0.3cm 0cm 0cm 0},clip]{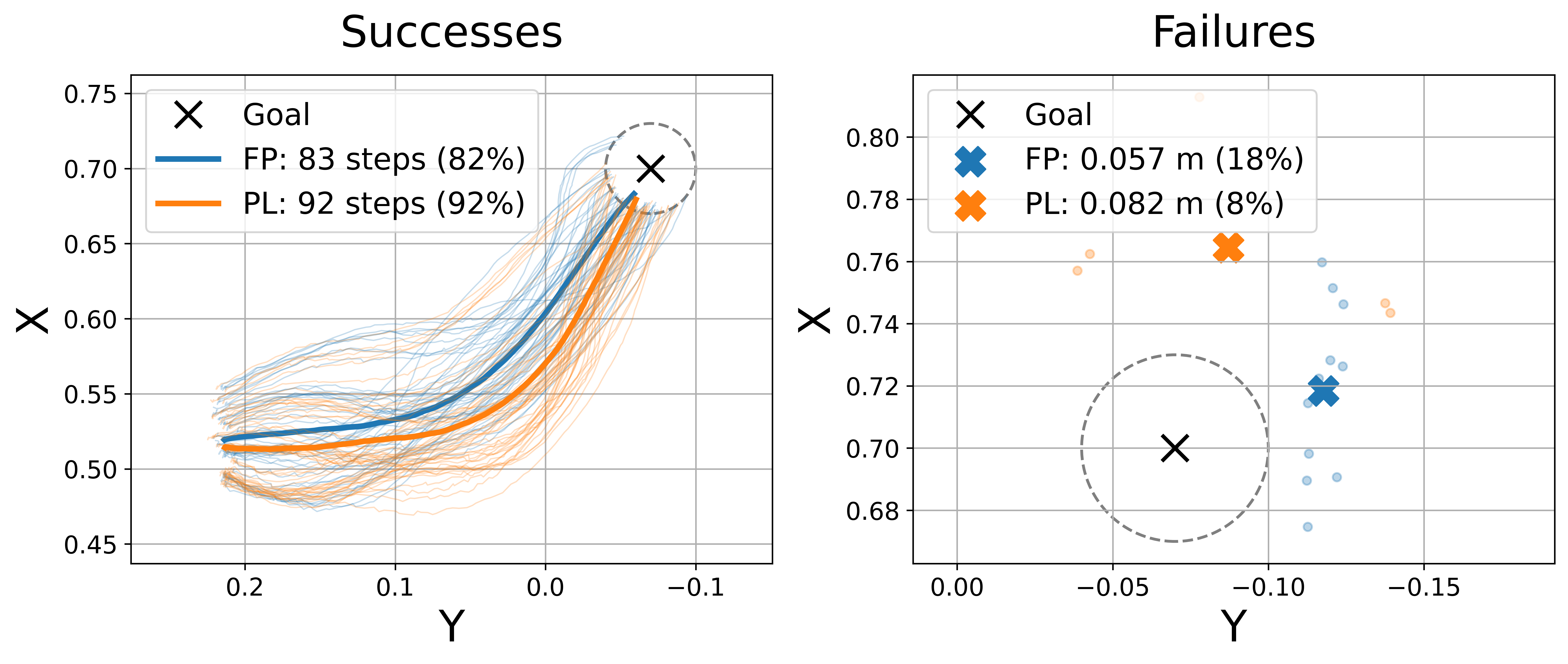} 
\vspace{-0.6cm}
\caption{Comparison of the best‐performing \textit{FP} and \textit{PL} policies on the pushing task without center of mass variation. Dashed circles denote the success threshold. \textbf{Left}. Successful box trajectories (faded lines) and their average ones (bold lines); legend values denote the average number of steps and the percentage of successful trials. \textbf{Right}. End box positions of failed trajectories (faded dots) and their centroids (bold crosses); legend values denote the average final distance to the goal and the percentage of failures.}
\label{fig:box_trajectories}
\vspace{-0.35cm}
\end{figure}

\subsection{Discussion}
\label{sec:eval_discussion}

\textbf{1) How does contextual information influence zero-shot generalization and sim-to-real transfer?}
Conditioning policies on contextual cues, either via a ground-truth oracle or a learned estimator, consistently yields substantial performance gains compared to a context-agnostic baseline.
We observe these benefits across both simulation and real-robot experiments: context-aware controllers achieve higher returns and success rates under domain shifts, with the gap widening as the context dimensionality increases.
While domain randomization alone offers some robustness, it fails to match the stability and peak performance of context-informed methods.
\textit{Oracle} appears to define an empirical upper bound, and no learned strategy surpasses it, particularly in high-dimensional settings where inferring accurate embeddings remains challenging.
Although previous works~\cite{contextualize_me,jcpl} have suggested that some forms of learned supervision may surpass this upper bound, our results do not support this conclusion.

\begin{figure}[H]
    \centering
    \subfigure[]{\includegraphics[width=0.45\textwidth,trim={4cm 0cm 1.5cm 0.1cm},clip]{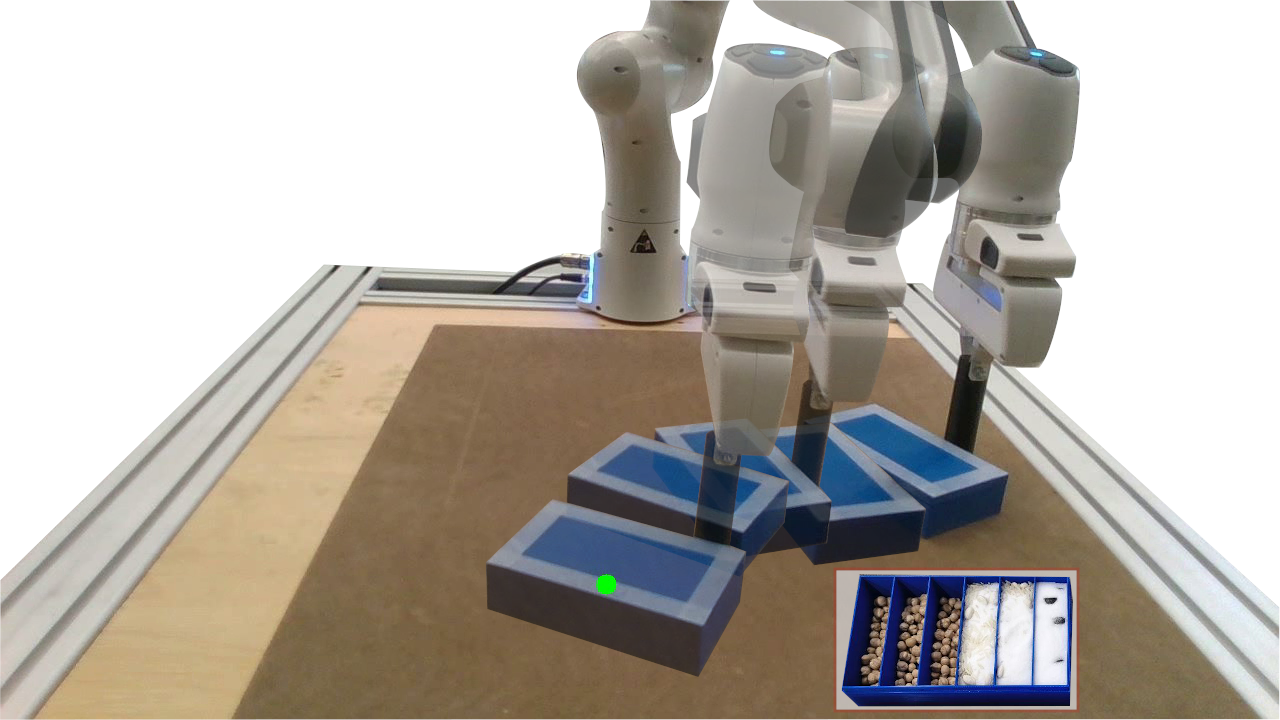}\label{fig:pushing_fp_left}}
    \hspace{1cm}
    \subfigure[]{\includegraphics[width=0.45\textwidth,trim={4cm 0cm 1.5cm 0.1cm},clip]{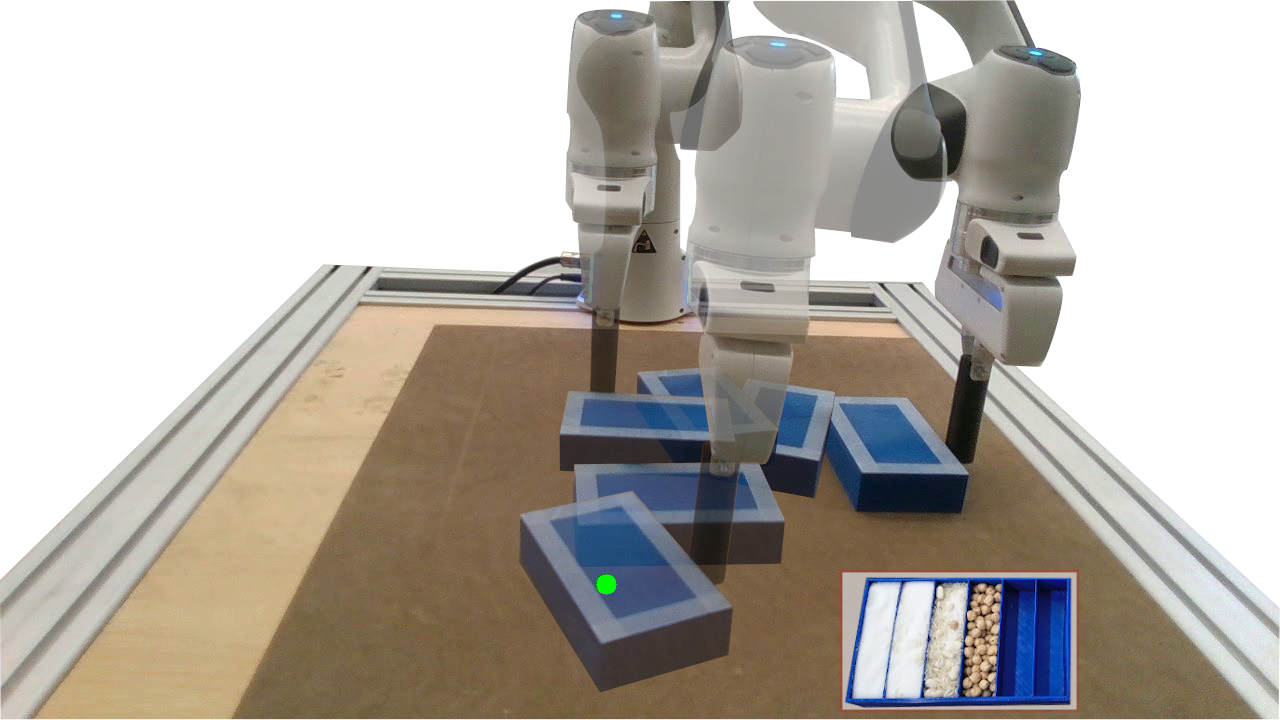}\label{fig:pushing_fp_right}}
    \caption{Representative real-robot pushing trajectories generated by the \textit{FP} policy for two different center-of-mass configurations. The green dot indicates the target goal position, while insets in the bottom-right corners visualize the internal contents of each box.}
    \label{fig:pushing_fp}
\end{figure}

\textbf{2) Which supervision strategy is most effective?}
Results suggest that the effectiveness of a supervision strategy depends on the considered task.
However, since the present analysis is based on only two task domains, the observed trends are not sufficient to establish general criteria linking task characteristics to the most suitable supervision strategy, and should therefore be interpreted as task-specific observations rather than broadly applicable guidelines.
This current evidence, therefore, underscores the need for further investigation to determine whether consistent correlations between task characteristics and supervision strategies can be identified.
Establishing such correlations would facilitate more principled method selection, reducing the need for exhaustive empirical comparisons, and potentially inform the design of a supervision strategy that outperforms the current baselines under all conditions.
Notably, policies trained using weak supervision through policy loss demonstrate competitive performance across both tasks, despite some variability.
Moreover, models that can selectively aggregate information across transitions, such as LSTMs and Transformers, consistently exhibit improved performance and reduced variance compared to feed-forward counterparts, with Transformers in particular emerging as the most reliable and high-performing architecture.

\textbf{3) What are the main challenges in evaluating generalization across varying contexts for sim-to-real?}
Assessing generalization across varying contexts requires distinct training, validation, and test sets, each sampled from the relevant context space.
However, this presents two key challenges: computational cost and limitations when testing on real hardware.
On the computational side, frequent evaluation of policies across the full validation set during training incurs significant time and resource overhead.
This is particularly problematic in high-dimensional contexts, where the number of required samples grows exponentially.
While reducing the evaluation frequency or using a sparser subset of contexts can alleviate the computational burden, it may lead to suboptimal model selection or misleading conclusions.
On the real robot, additional challenges arise due to the uncertainty in context parameters, such as friction, which are not easily measurable.
This can result in the testing of policies on a non-representative subset of the context space.
In response to this challenge, we recommend considering multiple context configurations when testing the policies on the real robot to ensure more reliable assessments of sim-to-real transfer.

\subsection{Limitations}
\label{sec:eval_limitations}

The present study is subject to limitations that should be considered when interpreting the results and their generalizability.
First, the analysis is restricted to an off-policy RL setting, as the considered training framework relies on replay-buffer transitions for context inference.
This restriction is particularly relevant for \textit{PL}, whose supervision signal is obtained by backpropagating the policy objective through the context estimator and is therefore inherently tied to off-policy training.
By contrast, \textit{GT} and \textit{FP} are trained through auxiliary objectives that are decoupled from the policy update and could, in principle, be adapted to on-policy algorithms by introducing a separate buffer for context-estimator training.
As a consequence, the conclusions drawn here should be understood as applying to the off-policy case, and do not establish whether the same observations would hold for on-policy methods.
A second limitation concerns the choice of the RL algorithm within the off-policy setting.
Since the experimental evaluation is restricted to SAC, it remains unclear to what extent the relative behavior of the supervision strategies would be preserved under other off-policy algorithms.
Therefore, the generalizability of the observed results to alternative RL algorithms and training settings remains to be investigated.
Finally, the study considers relatively low-dimensional context spaces, comprising at most four physical parameters.
While this setting is representative of many sim-to-real applications in which only a limited number of physical factors are expected to vary meaningfully, it does not establish how the proposed supervision strategies would scale to substantially higher-dimensional cases.


\section{Conclusion}
In this work, we examine the potential of context-aware policies to improve sim-to-real transfer in robotic control and analyze how different supervision strategies for learning contextual representations affect zero-shot generalization.
We conduct a systematic evaluation across simulated and real-world tasks using a Franka Emika Panda robot within an off-policy reinforcement learning setting based on SAC.
Experimental results demonstrate that conditioning policies on contextual information consistently enhances robustness to domain shifts, compared to standard domain randomization.
Among the evaluated approaches, weak supervision through policy loss achieves competitive performance across tasks.
Nonetheless, no single strategy consistently outperforms the others, indicating that the effectiveness of supervision methods may be task-specific.
However, given the limited number of task domains considered, the observed differences between supervision strategies do not yet support general conclusions about how supervision should be selected as a function of task characteristics.
We also highlight key challenges in evaluating generalization, particularly the computational burden of validation across high-dimensional context spaces and the difficulty of assessing performance on physical systems with unobservable parameters. Addressing these limitations is essential for reliable benchmarking and deployment.

Future research should aim to identify principled criteria for selecting appropriate supervision strategies based on task properties through broader evaluations across task domains and training settings.
In particular, it would be valuable to investigate the extent to which the observed results generalize to other off-policy reinforcement learning algorithms and to explore extensions of the framework to on-policy settings.
In addition, it would be valuable to explore alternative ways of leveraging privileged simulator information, both by considering different architectural choices (e.g., teacher-student) and by investigating supervision strategies that learn task-relevant latent representations from ground-truth signals rather than directly regressing physical parameters~\cite{privileged_info1,privileged_info2}.
Moreover, extending context-aware policies to explicitly account for uncertainty in context estimation represents a promising direction for improving robustness when the underlying dynamics cannot be reliably inferred from limited data~\cite{context_uncertainty}.
Finally, future work should also compare explicit context estimation with memory-based approaches, where recurrent policies infer dynamics through internal memory states derived from past observations~\cite{s2r_memory_based,pushing_bergmann}.

\section*{Declaration of Generative AI and AI-assisted technologies in the writing process}
During the preparation of this work, the authors used ChatGPT in order to improve readability and language. After using this tool/service, the authors reviewed and edited the content as needed and take full responsibility for the content of the publication.

\section*{Declaration of competing interest}
The authors affirm that there are no known competing interests or financial relationships that could be perceived as potential conflicts of interest.

\section*{Acknowledgement}
This work was supported in part by Industrial Graduate School Collaborative AI \& Robotics (CoAIRob), in part by the Swedish Knowledge Foundation under Grant Dnr:20190128, and the Knut and Alice Wallenberg Foundation through Wallenberg AI, Autonomous Systems and Software Program (WASP).

{\small

\bibliographystyle{elsarticle-num} 

\bibliography{references}
}

\newpage

\vfill
\end{document}